\documentclass[11pt,a4paper]{article}
\usepackage[hyperref]{acl2020}
\usepackage{url}
\usepackage{times}
\usepackage{helvet}
\usepackage{courier}
\usepackage{amsmath, amssymb}
\usepackage{bm}
\usepackage{graphicx}
\usepackage{subfig}
\usepackage{multirow}
\usepackage{multicol}
\usepackage{algorithm}
\usepackage[noend]{algpseudocode}
\usepackage{enumitem}
\usepackage{fancyvrb}
\usepackage{booktabs}
\usepackage{arydshln}
\usepackage{comment}
\usepackage{bm}
\usepackage{xcolor}
\usepackage{booktabs}
\usepackage{arydshln}
\usepackage{textcomp}

\newcommand{\bvec}[1]{\mbox{\boldmath $#1$}}

\usepackage{microtype}

\aclfinalcopy % Uncomment this line for the final submission
\def\aclpaperid{3248} %  Enter the acl Paper ID here

\newcommand{\CHECKI}[1]{{#1}}

\title{Single Model Ensemble using Pseudo-Tags and Distinct Vectors}

\author{Ryosuke Kuwabara \\
  The University of Tokyo  \\
  \texttt{\scriptsize kuwabara@nlab.ci.i.u-tokyo.ac.jp} \\\And
  Jun Suzuki \\
  Tohoku University, RIKEN  \\
  \texttt{\scriptsize jun.suzuki@ecei.tohoku.ac.jp} \\\And
  Hideki Nakayama \\
  The University of Tokyo  \\
  \texttt{\scriptsize nakayama@ci.i.u-tokyo.ac.jp}  \\}

\date{}

\begin{document}
\maketitle

\begin{abstract}
Model ensemble techniques often increase task performance in neural networks; however, they require increased time, memory, and management effort.
In this study, we propose a novel method that replicates the effects of a model ensemble with a single model.
Our approach creates $K$-virtual models within a single parameter space using $K$-distinct pseudo-tags and $K$-distinct vectors.
Experiments on text classification and sequence labeling tasks on several datasets demonstrate that our method emulates or outperforms a traditional model ensemble with $1/K$-times fewer parameters.
%
%Experiments on text classification and sequence labeling tasks demonstrate that our method not only emulates the performance of a normal ensemble in all tasks, but outperforms the normal ensemble in some text classification tasks with 89$\%$ less parameters.

\end{abstract}

\section{Introduction}

A model ensemble is a promising technique for increasing the performance of neural network models \cite{Hansen:1990:NNE:628297.628429, Krogh:1994:NNE:2998687.2998716}. 
This method combines the outputs of multiple models that are individually trained using the same training data. 
Recent submissions to natural language processing(NLP) competitions are primarily composed of neural network ensembles \cite{bojar-etal-2018-findings, barrault2019findings}. 
Despite its effectiveness, a model ensemble is costly.
Because it handles multiple models, it requires increased time for training and inference, increased memory, and greater management effort.
%%%
% A normal $K$-model ensemble occupies $K$ times the memory space than a single model.
%
%Furthermore, allocating multiple state-of-the-art models onto edge devices is not always realistic.
%
%For instance, BERT \cite{DBLP:journals/corr/abs-1810-04805} occupies about 1 GB memory. 
%
%Therefore, ensemble learning cannot always be applied to real use.
Therefore, the model ensemble technique cannot always be applied to real systems, as many systems, such as edge devices, must work with limited computational resources.
\begin{figure}[t]
  \centering
  \hspace{-0.41cm}
  \includegraphics[width=0.45\textwidth]{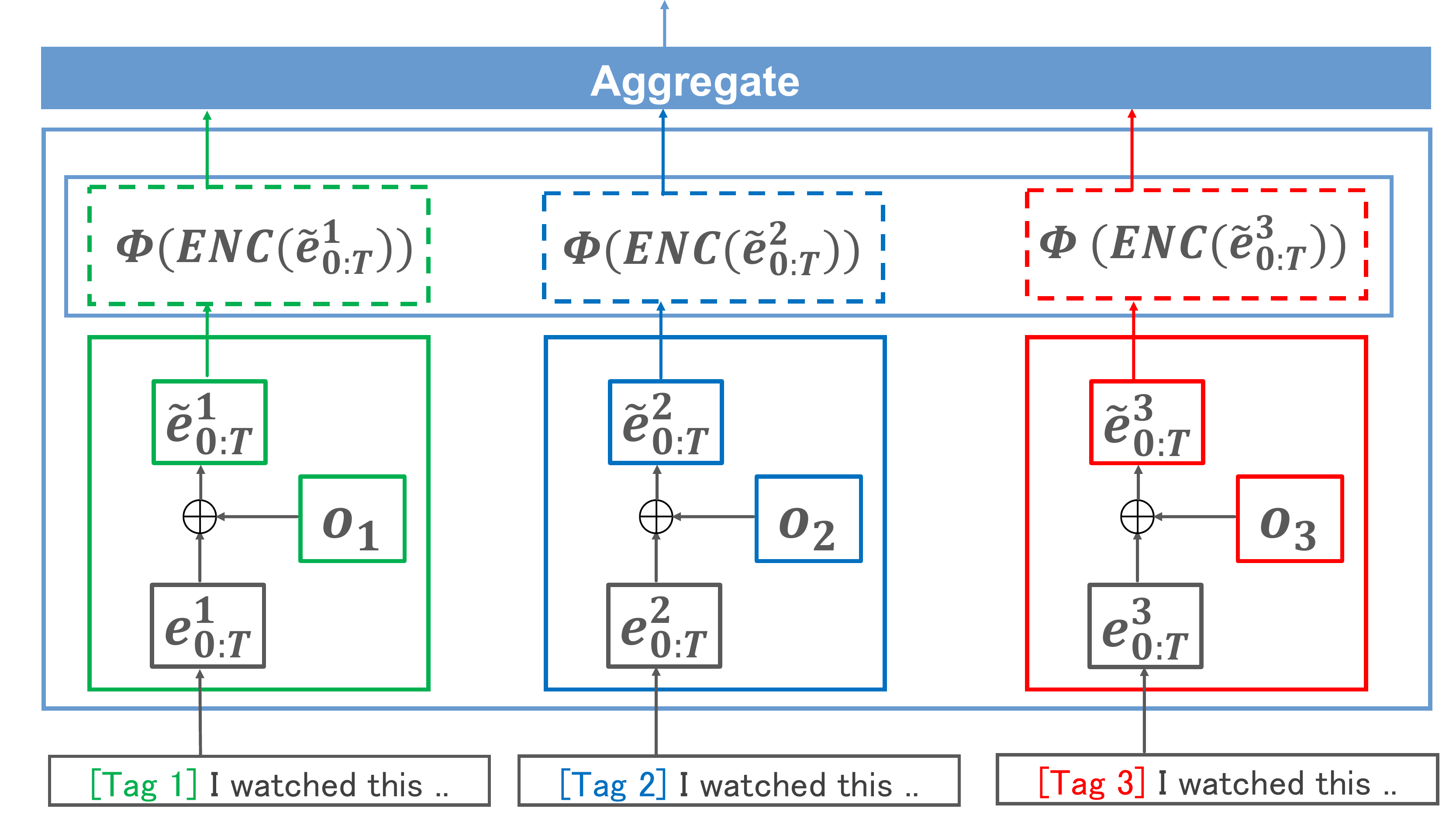}
  \caption{Overview of our proposed method. A single model processes the same input with distinct pseudo-tags. Each pseudo-tag defines the $k$-th virtual model, and the corresponding vector $\bvec{o}_{k}$ is added to the embedding. Thus, the model function of a singe model $\phi \left(\textsc{enc}(\cdot)\right)$ generates different outputs.}
  \label{fig:arch1}
  \vskip -2mm
\end{figure}

In this study, we propose a novel method that replicates the effects of the ensemble technique with a single model.
Following the principle that aggregating multiple models improves performance, we create multiple virtual models in a shared space.
%Our method consists of two parts 
%To alleviate the costs, we propose a new method that can imitate the effects of the ensemble technique with a single model. 
%
%In our method, the training data is virtually inflated $K$ times with $K$ distinct pseudo-tags appended to the data. 
Our method virtually inflates the training data $K$ times with $K$-distinct pseudo-tags appended to all input data. 
It also incorporates $K$-distinct vectors, which correspond to pseudo-tags.
%When processing time-series data, each pseudo tag $k \in K$ is attached at the beginning, and $k$-th vector is added to the embedding at each timestep of the input.
Each pseudo-tag $k \in \{1,\dots,K\}$ is attached to the beginning of the input sentence, and the $k$-th vector is added to the embedding vectors for all tokens in the input sentence.
Fig.~\ref{fig:arch1} presents a brief overview of our proposed method.
Intuitively, this operation allows the model to shift the embedding of the same data to the $k$-th designated subspace and can be interpreted as explicitly creating $K$ virtual models in a shared space.
We thus expect to obtain the same (or similar) effects as the ensemble technique composed of $K$ models with our $K$ virtual models generated from a single model. 

Experiments in text classification and sequence labeling tasks reveal that our method outperforms single models in all settings with the same parameter size.
Moreover, our technique emulates or surpasses the normal ensemble with $1/K$-times fewer parameters on several datasets.

\section{Related Work}\label{relatedwork}

The neural network ensemble is a widely studied method \cite{Hansen:1990:NNE:628297.628429, Krogh:1994:NNE:2998687.2998716, Hashem94optimallinear, Opitz1996ActivelySF}; however studies have focused mainly on improving performance while ignoring cost, such as computational cost, memory space, and management cost.

Several methods have overcome the shortcomings of traditional ensemble techniques. %ensemble learning \cite{Swann1998FastCL,DBLP:journals/corr/XieXZ13, DBLP:journals/corr/HuangLPLHW17, hinton2015distilling}.
For training Snapshot Ensembles, \cite{DBLP:journals/corr/HuangLPLHW17} used a single model to construct multiple models by converging into multiple local minima along the optimization path. 
For inference distillation, \cite{hinton2015distilling} transferred the knowledge of the ensemble model into a single model. 
These methods use multiple models either during training or inference, which partially solves the negative effects of the traditional ensemble.

%%%%%%%%%%%%
The incorporation of pseudo-tags is a standard technique widely used in the NLP community, ~\cite{sennrich-etal-2016-controlling,johnson-etal-2017-googles}.
However, to the best of our knowledge, our approach is the first attempt to incorporate pseudo-tags as an identification marker of virtual models within a single model.

%%%%%%%%%%%%
The most similar approach to ours is dropout \cite{JMLR:v15:srivastava14a},
which stochastically omits each hidden unit during each mini-batch, and in which all units are utilized for inference.
\citet{DBLP:journals/corr/HuangLPLHW17} interpreted this technique as implicitly using an exponential number of virtual models within the same network.
As opposed to dropout, our method explicitly utilizes virtual models with a shared parameter, which is as discussed in Section \ref{experiments}, complementary to dropout.

\section{Base Encoder Model} 

\CHECKI{ %%%
%\paragraph{Model Function}
%Our baseline model generally processes the sequence of tokens $\bvec{x}_{i:T}$ as inputs and yield $\bvec{y}_{i:T}$ as outputs, where $\bvec{x}_{t}$ and $\bvec{y}_{t}$ are the one-hot vectors of the $t$-th token in the input and output sequences, respectively.
%

The target tasks of this study are text classification and sequence labeling.
The input is a sequence of tokens (i.e., a sentence).
%
%Then, given a sentence, the model yields a sequence of tokens.
%
Here, $\bm{x}_{t}$ denotes the one-hot vector of the $t$-th token in the input.
%
%Moreover, we define $\bm{x}_{1:T}$ to represent the list of vectors $(\bm{x}_{1}, \bm{x}_{2}, \dots, \bm{x}_{T})$ that corresponds the input sentence, where $T$ is the number of tokens in the input.
%
Let $\bm{E} \in \mathbb{R}^{D \times |\mathcal{V}|}$ be the embedding matrices where $D$ is the dimension of the embedding vectors and $\mathcal{V}$ is the vocabulary of the input.

We obtain the embedding vector $\bm{e}_{t}$ at position $t$ by $\bm{e}_{t} = \bm{E}\bm{x}_{t}$. 
Here, we introduce the notation $\bm{e}_{1:T}$ to represent the list of vectors $(\bm{e}_{1}, \bm{e}_{2}, \dots, \bm{e}_{T})$ that correspond to the input sentence, where $T$ is the number of tokens in the input.
Given $\bm{e}_{1:T}$, the feature (or hidden) vectors $\bm{h}_{t}\in\mathbb{R}^{H} $ for all $t\in\{1,\dots,T\}$ are computed as an encoder neural network $\textsc{enc}(\cdot)$, where $H$ denotes the dimensions of the feature vector.
Namely, 
\begin{align}
\bm{h}_{1:T} = \textsc{enc} \left(\bm{e}_{1:T}\right).
\label{eq:encoder}
\end{align}
Finally, 
%let $\bm{W} \in \mathbb{R}^{C \times H}$ be a linear transformation matrix, where $C$ is the number of output classes.
%
the output $\widehat{\bm{y}}$ given input $\bm{x}_{1:T}$ is estimated as %follows:
$\widehat{\bm{y}} = \phi \left(\bvec{h}_{1:T}\right)$ 
where $\phi\left(\cdot \right)$ represents the task dependent function (e.g., a softmax function for text classification and  a conditional random field layer for sequence labeling). 
It should be noted that the form of the output $\widehat{\bm{y}}$ differs depending on the target task.
%
% For example, $\widehat{\bm{y}}$ is an estimated distribution or a list of estimated distributions (e.g., $\widehat{\bm{y}}=\widehat{\bm{y}}_{1:T}$) of output classes preliminarily defined by text classification or sequence labeling tasks, respectively.

%
} %%%%%

\CHECKI{
\section{Single Model Ensemble using Pseudo-Tags and Distinct Vectors}\label{method}

In this section, we introduce the proposed method, which we refer to as \textsc{SingleEns}. 
Fig.~\ref{fig:arch1} presents an overview of the method. 
The main principle of this approach is to create different virtual models within a single model. 

We incorporate \textbf{pseudo-tags} and \textbf{predefined distinct vectors}. % that correspond to each pseudo tag. 
%
%Each pseudo tag and vector define a virtual model.
%
For the pseudo-tags, we add special tokens $\{\bm{\ell}_k\}^{K}_{k=1}$ to the input vocabulary, where hyper-parameter $K$ represents the number of virtual models. 
For the predefined distinct vectors, we leverage mutually orthogonal vectors $\{\bm{o}_k\}^{K}_{k=1}$, where the orthogonality condition requires satisfying $\bvec{o}_{k} \cdot \bvec{o}_{k'} \simeq 0$ for all $(k,k')$ when $k\ne k'$.

Finally,
we assume that all input sentences start from one of the pseudo-tags.
%attach a pseudo tag, e.g., $\bm{\ell}_k$, to the beginning of the input $\bm{x}_{1:T}$.
%
We then add the corresponding orthogonal vector $\bm{o}_{k}$ of the attached pseudo-tag $\bm{\ell}_k$ to the embedding vectors at all positions. 
%%
%\begin{align}
%\bm{o}_{k} \in \bm{O},
%\end{align}
%where $k$ represents $k$-th virtual model and $\bvec{O}$ is a set of pre-defined orthogonal vectors.
%
The new embedding vector $\tilde{\bm{e}}_{0:T}$ is written in the following form:
%
%Therefore, the model function equipped with the our method is described as follows:
%
%
\begin{align}
\tilde{\bm{e}}^{(k)}_{0:T} = (\bm{\ell}_k, \bm{e}_{1}+\bvec{o}_{k}, \bm{e}_{2}+\bvec{o}_{k}, \dots, \bm{e}_{T}+\bvec{o}_{k}).
%\bvec{y}_{t_{k}} &= f \left(\bvec{W}\bvec{h}_{t_{k}}\right), \\
%
%\bvec{h}_{t_{k}} &= \textsc{enc} \left(\ell_k, \tilde{\bm{e}}_{t}+\bvec{o}_{k}\right), \\
%
%\text{s.t.} \indent &  \bvec{o}_{k} \cdot \bvec{o}_{j \neq k} \simeq 0\nonumber
%
\end{align}
We substitute ${\bm{e}}_{1:T}$ in Eq.~\ref{eq:encoder} by $\tilde{\bm{e}}^{(k)}_{0:T}$ in the proposed method.

An intuitive explanation of the role of pseudo-tags is to allow a single model to explicitly recognize differences in homogeneous input, while the purpose of orthogonal vectors is to linearly shift the embedding to the virtual model's designated direction.
Therefore, by combining these elements, we believe that we can define virtual models within a single model and effectively use  the local space for each virtual model. Aggregating these virtual models can then  result in imitation of ensemble.
} %%% TODO

\begin{table}
    \centering
    \footnotesize
    \tabcolsep=2pt
    \begin{tabular}{l|c|lrl}
    \hline 
    {Dataset} & {Model} & {Method}  & {{\#} params} & {Accuracy}  \\ 
    \hline\hline
     && \textsc{Single} &   12 M & \,\,87.03 \,\hspace*{10mm}\\
                      \cdashline{3-5}[3pt/1pt]
    &\multirow{1}{*}{\textsc{Tfm:}}& \textsc{1/K Ens} & 14 M  & \,\,81.93 ($-5.10$)\\
     & \multirow{1}{*}{\textsc{GloVe}} & \textbf{\textsc{SingleEns}}       &   12 M & \,\,{87.30} ($+0.27$)\\
       %  \cdashline{3-5}[3pt/1pt]
    \ IMDB            && \textsc{NormalEns}      & 108 M  & \,\,\bf 87.67 ($+0.64$)\\ 
    \cline{2-5}
     && \textsc{Single} & 400 M  & \,\,91.99 \,\hspace*{10mm}\\
    \cdashline{3-5}[3pt/1pt]
    &\multirow{1}{*}{\textsc{Tfm:}}& \textsc{1/K Ens} & 1000 M  & \,\,90.63 ($-1.36$) \\
      &\multirow{1}{*}{\textsc{BERT}} & \textbf{\textsc{SingleEns}}       & 400 M  & \,\,\textbf{92.91} ($+0.92$) \\
        %\cdashline{3-5}[3pt/1pt]
    \             && \textsc{NormalEns}      & 3600 M & \,\,92.75 ($+0.76$) \\ 
    \hline
     && \textsc{Single} & 400 M  & \,\,81.75 \,\hspace*{10mm}\\
        \cdashline{3-5}[3pt/1pt]
      &\multirow{1}{*}{\textsc{Tfm:}}& \textsc{1/K Ens} & 1000 M  & \,\,82.67 ($+0.92$)\\
     Rotten &\multirow{1}{*}{\textsc{BERT}} & \textbf{\textsc{SingleEns}}       & 400 M  & \,\,\textbf{85.01} ($+3.26$) \\
    \          && \textsc{NormalEns}      & 3600 M & \,\,82.57 ($+0.82$) \\ 
    \hline
     && \textsc{Single} & 400 M  & \,\,87.18 \,\hspace*{10mm}\\
        \cdashline{3-5}[3pt/1pt]
      &\multirow{1}{*}{\textsc{Tfm:}}& \textsc{1/K Ens} & 1000 M  & \,\,80.27 ($-6.91$)  \\
     RCV1 &\multirow{1}{*}{\textsc{BERT}} & \textbf{\textsc{SingleEns}}       & 400 M  & \,\,89.16 ($+1.98$) \\
    \             && \textsc{NormalEns}      & 3600 M & \,\,\textbf{90.01} ($+2.83$)\\ 
    \hline
    \end{tabular}
    \caption{Test accuracy and parameter size for text classification tasks. Our method, \textbf{\textsc{SingleEns}}, outperformed \textsc{Single} and \textsc{1/K Ens} on all datasets. Most notably, \textbf{\textsc{SingleEns}} surpassed  \textsc{NormalEns} on IMDB and Rotten with 1/9 fewer parameters.}
    \label{table:table1}
    \vskip -2mm
\end{table}

\section{Experiments}\label{experiments}
To evaluate the effectiveness of our method, we conducted experiments on two tasks: text classification and sequence labeling.
We used the IMDB \cite{maas-EtAl:2011:ACL-HLT2011}, Rotten \cite{pang-lee-2005-seeing}, and RCV1 \cite{Lewis:2004:RNB:1005332.1005345} datasets for text classification, and the CoNLL-2003 \cite{tjong-kim-sang-de-meulder-2003-introduction} and CoNLL-2000 datasets \cite{tjong-kim-sang-buchholz-2000-introduction} for sequence labeling.

We used the Transformer model~\cite{DBLP:journals/corr/VaswaniSPUJGKP17} as the base model for all experiments, and its token vector representations were then empowered by pretrained vectors of GloVe,  \cite{pennington-etal-2014-glove}, BERT \cite{DBLP:journals/corr/abs-1810-04805}, or ELMo \cite{peters-etal-2018-deep}.
The models are referred to as \textsc{Tfm:GloVe},  \textsc{Tfm:BERT}, and \textsc{Tfm:ELMo}, respectively.%
\footnote{See Appendix \ref{subsec:appendix1} for detailed experimental settings.}
For \textsc{Tfm:BERT}, we incorporated the feature (or hidden) vectors of the final layer in the BERT model as the embedding vectors while adopting drop-net technique~\cite{Zhu2020Incorporating}.
%
% Implementation details are provided in a report by ~\cite{Zhu2020Incorporating}.
%
All the models have dropout layers to assess the complementarity of our method and dropout.

We compared our method (\textbf{\textsc{SingleEns}}) to a single model (\textbf{\textsc{Single}}), a normal ensemble (\textbf{\textsc{NormalEns}}), and a normal ensemble in which each component has approximately $1/K$ parameters%
\footnote{Because BERT requires a fixed number of parameters, we did not reduce the parameters accurately for \textsc{1/K} \textsc{Tfm:BERT}.}
(\textbf{\textsc{1/K Ens}}).\footnote{See Appendix \ref{subsec:appendix1} for detailed experimental settings.}
Although other ensemble-like methods discussed in Section \ref{relatedwork} could have been compared (e.g., snapshot ensemble, knowledge distillation, or dropout during testing to generate predictions and aggregate them), they are imitations of a normal ensemble, and we assumed that the results of a normal ensemble were upper-bound.
We used $K=9$ for reporting the primary results of \textsc{NormalEns}, \textsc{1/K Ens}, and \textsc{SingleEns}.
We thus prepared nine pseudo-tags $\{\bm{\ell}_k\}^{9}_{k=1}$ in the same training (trainable) and initialization manner as other embeddings. We created untrainable distinct vectors $\{\bm{o}_k\}^{9}_{k=1}$ using the implementation by ~\citet{saxe2013exact} that was prepared in PyTorch's default function, \texttt{torch.nn.init.orthogonal}.
We empirically determined the correct scaling for the distinct vectors as 1 out of 1, 3, 5, 10, 30, 50, 100, and the scale that was closest to the model's embedding vectors.
We obtained the final predictions of $K$ ensemble models by averaging and voting the outputs of individual models for text classification and sequence labeling, respectively.
The results were obtained by the averaging five distinct runs with different random seeds.

\begin{table}
    \centering
    \footnotesize
    \tabcolsep=2pt
    \begin{tabular}{l|c|lrl}\hline 
    {Dataset} & {Model} & {Method}  & {{\#} params} & {F1 Score}  \\ 
    \hline\hline
      && \textsc{Single} & 100 M   & \,\,91.93 \,\hspace*{2mm}\\
     \cdashline{3-5}[3pt/1pt]
   CoNLL   &\multirow{1}{*}{\textsc{Tfm:}}& \textsc{1/K ENS} & 150 M  & \,\,91.65 ($-0.28$) \\
    2003    &\multirow{1}{*}{\textsc{ELMo}}   & \textbf{\textsc{SingleEns}}          & 100 M   & \,\,92.37 ($+0.44$) \\
    \              &\textsc{}& \textsc{NormalEns}     & 900 M  & \,\,\textbf{92.86} ($+0.93$) \\ \hline
    && \textsc{Single} & 100 M   & \,\,96.42 \,\hspace*{2mm}\\\cdashline{3-5}[3pt/1pt]
    CoNLL   &\textsc{Tfm:}& \textsc{1/K ENS} & 150 M  & \,\,95.67 ($-0.75$) \\
    2000    &\multirow{1}{*}{\textsc{ELMo}}   & \textbf{\textsc{SingleEns}}          & 100 M   & \,\,96.56 ($+0.14$) \\
    &\textsc{}& \textsc{NormalEns} & 900 M   & \,\,\textbf{96.67} ($+0.25$) \hspace*{2mm}\\\hline
    \end{tabular}
    \caption{Test F1 score and parameter size for sequence labeling tasks.  Similarly to \textsc{NormalEns}, \textbf{\textsc{SingleEns}} improved the score even at high performance levels.}
    \label{table:table2}
    \vskip -2mm
\end{table}

\subsection{Evaluation of text classification}
\paragraph{Data}
We followed the settings used in the implementation by \citet{DBLP:journals/corr/abs-1810-05788} for data partition.%
\footnote{See Appendix \ref{subsec:appendix2} for data statistics.}
Our method, \textsc{SingleEns} inflates the training data by $K$ times. 
During the inflation, the $k$-th subset is sampled by bootstrapping \cite{EfroTibs93} with the corresponding $k$-th pseudo-tag.
%
%For \textsc{NormalEns} we attempted both bootstrapping and normal sampling and higher score was reported while only normal sampling was applied for \textsc{1/K Ens}.
For \textsc{NormalEns} and \textsc{1/K Ens}, we attempted both bootstrapping and normal sampling, and a higher score was reported.
%

%\subsubsection{Results}
\paragraph{Results}
Table \ref{table:table1} presents the overall results evaluated in terms of accuracy.
For both \textsc{Tfm:GloVe} and \textsc{Tfm:BERT}, \textsc{SingleEns} outperformed \textsc{Single} with the same parameter size. 
In our experiments, \textsc{SingleEns} achieved the best scores on IMDB and Rotten with \textsc{Tfm:BERT}; it recorded $92.91\%$ and $85.01\%$, which was higher than \textsc{NormalEns} by $0.16$ and $2.44$, respectively with 89$\%$ fewer parameters. 
The standard deviation of the results for the IMDB dataset was, 0.69 and 0.14 for \textsc{Single} and \textsc{SingleEns}, respectively, for \textsc{Tfm:GloVe}, and 0.34 and 0.11, respectively, for \textsc{Tfm:BERT}.
These results support the claim that explicit operations for defining $K$ virtual models have a significant effect for a single model and are complementary to normal dropout.
Through the series of experiments, we observed that the number of iterations of \textsc{SingleEns} was 1.0 \textasciitilde 1.5 times greater than that of \textsc{Single}.

\begin{figure*}[htpb]
\scalebox{1.0}{
    \begin{tabular}{cc}\hspace{-0.3cm}
      \begin{minipage}{0.48\hsize}\vspace{0.5cm}\hspace{-0.7cm}
        % \centering
         \includegraphics[scale=0.72, angle=0]{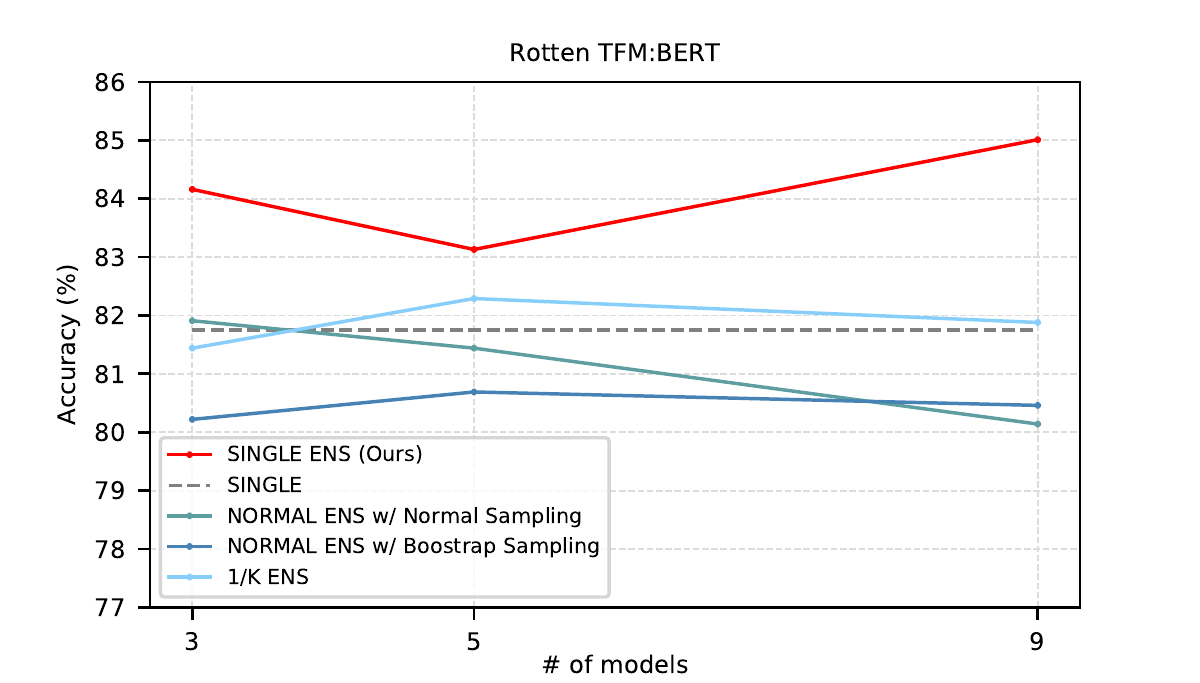}
        \caption{Accuracy depending on the number of models for each ensemble method on the Rotten dataset.}\label{fig:fig2}\hspace{-1.2cm}
      \end{minipage} & 
    %   \hspace{0.5cm}
      \begin{minipage}{0.48\hsize}\hspace{-0.6cm}
        % \centering
        \includegraphics[scale=0.75, angle=0]{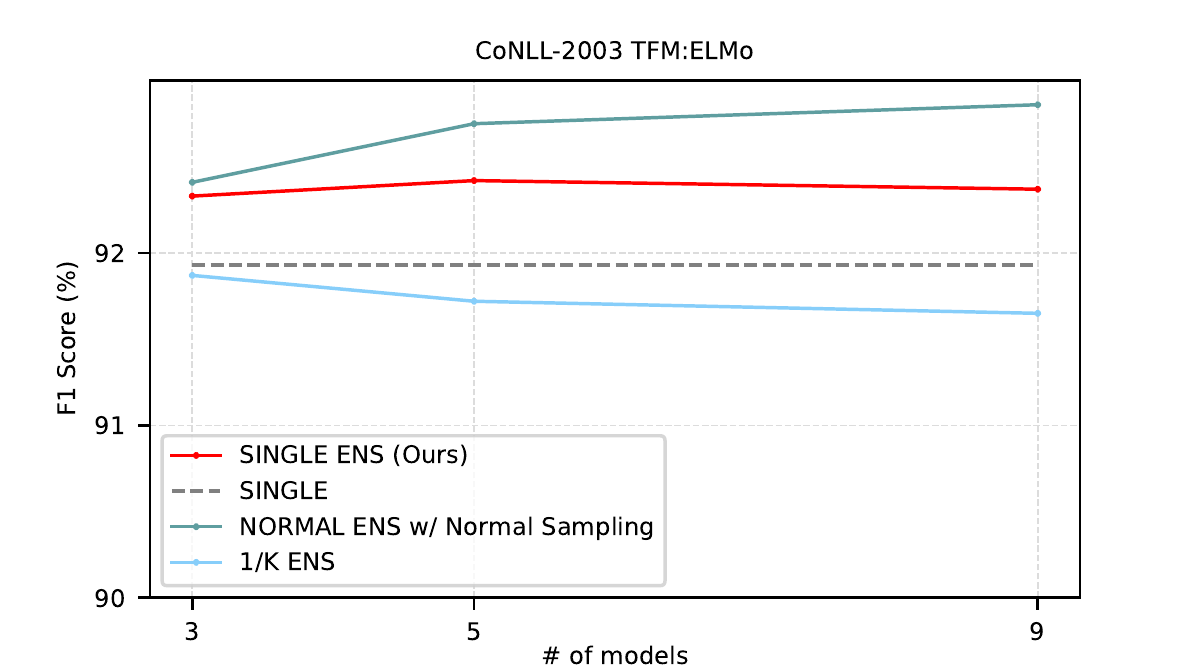}
        \caption{F1 score depending on the number of models for each ensemble method on CoNLL-2003.}
        \label{fig:fig3}
      \end{minipage}    
      \end{tabular}}
    \vskip -4mm
\end{figure*}

\subsection{Evaluation of sequence labeling}
\paragraph{Data}
We followed the instructions of the task settings used in CoNLL-2000 and CoNLL-2003.\footnote{The statistics of the datasets are presented in Appendix \ref{subsec:appendix2}.}
We inflated the training data by nine times for \textsc{SingleEns}, and normal sampling was used for \textsc{NormalEns} and \textsc{1/K Ens}.
Because bootstrapping was not effective for the task, the results were omitted.

\paragraph{Results}
As displayed in Table \ref{table:table2}, \textsc{SingleEns} surpassed \textsc{Single} by 0.44 and 0.14 on CoNLL-2003 and CoNLL-2000, respectively, for \textsc{TFM:ELMo} with the same parameter size. However, \textsc{NormalEns} produced the best results in this setting.
%
% The experimental results thus demonstrate that like the normal ensemble method, our method was effective in the high-performance range.
%
The standard deviations of the single model and our methods were 0.08 and 0.05, respectively, on CoNLL-2000.
Through the series of experiments, we observed that the number of iterations of \textsc{SingleEns} was 1.0 \textasciitilde 1.5 times greater than that of \textsc{Single}.

\begin{table}
    \centering
    \footnotesize
    \tabcolsep=4pt
        \begin{tabular}{lcc}
        \hline \   & IMDB & CoNLL-2003  \\ 
               Setting                & Accuracy %(\mathbf{\%}) 
                               & F1 Score %(\mathbf{\%}) 
                               \\ \hline
        \textsc{Single} & 91.99 & 91.93 \\
        1) Only pseudo-tags & 89.84  & 92.20 \\
        2) Random distinct vectors & 92.06 & 92.21  \\ 
        3) Random noise & 92.38  &  92.32\\
        \textsc{SingleEns}   & \textbf{92.91}  & \textbf{92.37}  \\ \hline
        \end{tabular}
    \caption{Comparison of proposed method (pseudo-tags + corresponding distinct vectors) with other settings. Pseudo-tags and distinct vectors appear to complement each other.}
    \label{table:effective methods}
    %\vskip -2mm
\end{table}

\begin{table}
    \centering
    \footnotesize
    \tabcolsep=4pt
        \begin{tabular}{lcc}
                \hline \   & IMDB & CoNLL-2003  \\ 
               Setting                & Accuracy %(\mathbf{\%}) 
                               & F1 Score %(\mathbf{\%}) 
                                \\ \hline
        \textsc{Single} & 91.99 & 91.93 \\
        1) Emb (\textsc{SingleEns}) & \textbf{92.91} & 92.37  \\
        2) Hidden & 90.68 & \textbf{92.45}  \\
        1) + 2) & 92.64 & 92.19  \\ \hline
        \end{tabular}
    \caption{Test metrics on IMDB and CoNLL-2003 with the pattern of three vector addition operations. Adding distinct vectors to only embeddings is the best or second best approach.}
    \label{table:operations}
   % \vskip -2mm
\end{table}

%%% %%%
\section{Analysis}\label{analysis}
In this section, we investigate the properties of our proposed method.
Unless otherwise specified, we use \textsc{Tfm:BERT} and \textsc{Tfm:ELMo} on IMDB and CoNLL-2003 for the analysis.

\paragraph{Significance of pseudo-tags and distinct vectors}
To assess the significance of using both pseudo-tags and distinct vectors, we conducted an ablation study of our method, \textsc{SingleEns}.
We compared our method with the following three settings:  1) Only pseudo-tags, 2) Random distinct vectors, and 3) Random noise.
In detail, the first setting (Only pseudo-tags) attached the pseudo-tags to the input without adding the corresponding distinct vectors. %This method was used to evaluate the effectiveness of using both pseudo-tags and distinct vectors. 
%
%If appending pseudo-tags is sufficient, the performance of our proposed method should not be higher than that of this method.
The second setting (Random distinct vectors) randomly shuffles the correspondence between the distinct vectors and pseudo-tags in every iteration during the training.
%
%If the performance of this method is as high as that of our proposed method, then our method can be considered a data augmentation method rather than a method with explicitly defined virtual models. 
%
%We conducted these two settings to evaluate the effectiveness of using only pseudo-tags or distinct vectors, which are both incorporated in \textsc{SingleEns}. 
%
Additionally, the third setting (Random noise) adds random vectors as the replacement of the distinct vectors to clarify whether the effect of incorporating distinct vectors is essentially identical to the random noise injection techniques or explicit definition of virtual models in a single model.  %The third method adds random noise, which we consider a substitute for distinct vectors. 
%
%If the proposed method is effective due to the noise, its performance should be lower because it feeds a fixed pattern of vectors while random noise feeds an exponential number of different vectors.
%

Table \ref{table:effective methods} shows the results of the ablation study.
This table indicates that using both pseudo-tags and distinct vectors, which matches the setting of \textsc{SingleEns}, leads to the best performance, while the effect is limited or negative if we use pseudo-tags alone or distinct vectors and pseudo-tags without correspondence. 
%Table \ref{table:effective methods} indicates that using both pseudo-tags and distinct vectors leads to the best performance, whereas the effect is small or negative if only one or the other is used. 
%
Thus, this observation explains that the increase in performance can be attributed to the combinatorial use of pseudo-tags and distinct vectors, and not merely data augmentation. % or random noise. %Therefore, the increase in performance can be attributed to the use of both pseudo-tags and distinct vectors, and not simply data augmentation or random noise.

We can also observe from Table \ref{table:effective methods} that the performance of \textsc{SingleEns} was higher than that of 3) Random noise.
Note that the additional vectors by \textsc{SingleEns} are fixed in a small number $K$ while those by Random noise are a large number of different vectors.
Therefore, this observation supports our claim that the explicit definition of virtual models by distinct vectors has substantial positive effects that are mostly irrelevant to the effect of the random noise.
This observation also supports the assumption that \textsc{SingleEns} is complementary to dropout.
Dropout randomly uses sub-networks by stochastically omitting each hidden unit, which can be interpreted as a variant of Random noise.
Moreover, it has no specific operations to define an explicitly prepared number of virtual models as \textsc{SingleEns} has. %In contrast, our method explicitly determines virtual models by pseudo-tags (virtual model definition) and distinct vectors (linear shift to subspace).
We conjecture that this difference yields the complementarity that our proposed method and dropout can co-exist.
\paragraph{Vector addition}
We investigated the patterns with which distinct vectors should be added: 1) Emb, 2) Hidden, and 3) Emb + Hidden.
Emb adds distinct vectors only to the embedding, while Hidden adds distinct vectors only to the final feature vectors. Emb + Hidden adds distinct vectors to both the embedding and final feature vectors.
As illustrated in Table \ref{table:operations}, adding vectors to the embedding is sufficient for improving performance, while adding vectors to hidden vectors has as adverse effect. 
This observation can be explained by the architecture of Transformer. 
The distinct vectors in the embedding are recursively propagated through the entire network without being absorbed as non-essential information since the Transformer employs residual connections~\cite{DBLP:journals/corr/HeZRS15}. %Because Transformer employs residual connections \cite{DBLP:journals/corr/HeZRS15}, a slight difference in the embedding recursively propagates through the entire network without being absorbed as non-essential information. 

\paragraph{Comparison with normal ensembles}
To evaluate the behavior of our method, we examined the relationship between the performance and the number of models used for training. Our experiments revealed that having more than nine models did not result in significant performance improvement; thus, we only assessed the results up to nine models.
Figs \ref{fig:fig2} and \ref{fig:fig3} present the metrics on Rotten and CoNLL-2003, respectively. 
The performance of our method increased with the number of models, which is a general feature of normal ensemble.
Notably, on Rotten, the accuracy of our method rose while that of other methods did not.
Investigation of this behavior is left for future work.

\section{Conclusion}
In this paper, we propose a single model ensemble technique called \textsc{SingleEns}.
The principle of \textsc{SingleEns} is to explicitly create multiple virtual models in a single model.
Our experiments demonstrated that the proposed method outperformed single models in both text classification and sequence labeling tasks.
Moreover, our method with \textsc{Tfm:BERT} surpassed the normal ensemble on the IMDB and Rotten datasets, while its parameter size was $1/K$-times smaller.
The results thus indicate that explicitly creating virtual models within a single model improves performance.
The proposed method is not limited to the two aforementioned tasks, but can be applied to any NLP as well as other tasks such as machine translation and image recognition.
Further theoretical analysis can also be performed to elucidate the mechanisms of the proposed method.

\paragraph{Acknowledgment}
The research results were achieved by ”Research and Development of Deep Learning Technology for Advanced Multilingual Speech Translation”, the Commissioned Research of National Institute of Information and Communications Technology (NICT), JAPAN. The work was partly supported by JSPS KAKENHI Grant Number 19H04162. We would like to thank Motoki Sato of Preferred Networks and Shun Kiyono of RIKEN for cooperating in preparing the experimental data.  We would also like to thank the three anonymous reviewers for their insightful comments.

\bibliography{acl2020}
\bibliographystyle{acl_natbib}

\clearpage
\onecolumn
\appendix

\section{Hyper-parameters and Ensemble Strategy}\label{subsec:appendix1}
% Detailed configuration of the experimental settings. Note that for \textsc{Tfm:ELMo},  we add Linear $\rightarrow$ Relu $\rightarrow$ LayerNorm between embedding and self-attention. The orthogonal vectors are added to the input of the self-attention with dropout.

% \begin{table}[h]
%     \centering
%     % \small
%     % \tabcolsep=4pt
%     \begin{tabular}{l|rrr}
%     \hline
%     Method & Accuracy & Sensitivity & Specificity \\ \hline
%     BoW + SVM & 72.6 \% & \textbf{93.5\%} &  43.1\% \\
%     BERT & \textbf{84.0\%} & 90.3\% &  \textbf{75.0\%} \\
%     \hline
%     \end{tabular}
%     \label{table:table5}
    
% \end{table}

\begin{table*}[h!]
    \centering
    \small
    \tabcolsep=4pt
    \begin{tabular}{l|rrr}
    \hline
     & \multicolumn{2}{c}{Text Classification} & \multicolumn{1}{c}{Sequence Labeling}\\ \hline
     & \textsc{Tfm:GloVe} & \textsc{Tfm:BERT}  & \textsc{Tfm:ELMo}  \\\hline\hline
    Embedding dimension & 200 & 768 &  256 \\
    Hidden dimension & 200 & 768 &  256 \\
    Number of layers & 6 & 6 &  6 \\
    Number of attention heads & 8 & 8 &  8 \\
    Frozen vectors & GloVe 200  &   BERT-Large &  ELMo 1024  \\\hline
            & - & - & 0.5(Emb)\\
            & 0.2 (Residual) & 0.5 (Residual)  & 0.2 (Residual)  \\
    Dropout      & 0.1 (Attention) & -   & 0.1 (Attention) \\
            &   -             &  -  & 0.1  (FF)    \\ \hline
    Label smoothing & 0.1& 0.1  & - \\ \hline
    Optimizer  & Adam & Adam & Adam  \\
    Initial learning rate & 0.0001 & 0.0001 & 0.0001 \\
    Batch size   & 64 & 128 & 32 \\
    Gradient Clipping  & 1.0 & 1.0  & 5.0 \\ \hline
    Aggrgation Strategy  & Averaging & Averaging & Voting \\
    Sampling strategy  & Normal $\&$ Bootstrapping & Normal $\&$ Bootstrapping & Normal \\ \hline
    \end{tabular}
    \caption{Hyper-parameters and ensemble strategies for \textsc{Single}, \textsc{NormalEns} and \textsc{SingleEns}. For \textsc{Tfm:BERT}, we followed the model architecture of \citet{Zhu2020Incorporating}. For \textsc{Tfm:ELMo} on sequence labeling, we referenced the architecture of   \citet{peters-etal-2018-deep} with replacing the encoder with Transformer. It should be noted that for \textsc{Tfm:ELMo},  we add Linear $\rightarrow$ Relu $\rightarrow$ LayerNorm between embedding and self-attention.}
    \label{table:table5}
\end{table*}

\begin{table*}[h!]
    \centering
    \small
    \tabcolsep=4pt
    \begin{tabular}{l|ll|r}
    \hline
     & \multicolumn{2}{c}{Text Classification} & \multicolumn{1}{c}{Sequence Labeling}\\ \hline
     & \textsc{Tfm:GloVe} & \textsc{Tfm:BERT}  & \textsc{Tfm:ELMo}  \\\hline\hline
    Embedding dimension & 50 & 64 &  370 \\
    Hidden dimension & 50 & 64 &  128 \\
    Frozen vectors & GloVe 50  &  BERT-Base & ELMo 256 \\\hline 
    Number of layers & 3  &   3 &  4   \\ 
    Number of attention heads & 10  &   8 &  8 \\
    Feed forward dimension & 128 & 128 &  128\\ \hline
    Aggregation Strategy  & Averaging & Averaging  & Voting \\
    Sampling strategy  & Normal & Normal  & Normal \\ \hline
    \end{tabular}
    \caption{Hyper-parameters and ensemble strategies for \textsc{1/K Ens}. The other values are same as Table \ref{table:table5}. It should be noted that we ensemble $K$ models of each sub model for final prediction.}
\end{table*}

\section{Data Statistics}\label{subsec:appendix2}

\begin{table*}[h!]
    \centering
    \small
    \tabcolsep=4pt
    \begin{tabular}{l|l|lrr}
    \hline
    Task & Dataset & Train  & Valid & Test  \\\hline\hline
                        & IMDB & 21,246 & 3,754 & 25,000 \\
    Text Classification & Rotten & 8,636 & 960 & 1,066 \\
                         & RCV 1 & 14,007 & 1,557 & 49,838 \\\hline
    Sequence Labeling & CoNLL-2003 & 14,987 & 3,466 & 3,684 \\\
                    & CoNLL-2000 & 8,926 & 2,012 & 2,012 \\\hline
    \end{tabular}
    \caption{Summary of the datasets. The values are the number of sentences contained in each dataset.}
\end{table*}
\end{document}